\documentclass[11pt]{article} 
\usepackage{aaai18}
\usepackage{color}
\usepackage{comment}
\usepackage{times}  
\usepackage{helvet}  
\usepackage{courier}  
\usepackage{url}  
\usepackage{graphicx}  
\graphicspath{ {./images/} } 
\usepackage{algorithm}
\usepackage[noend]{algpseudocode}
\usepackage[margin=.535in]{geometry}
\nocopyright
\usepackage[authoryear]{natbib} 

\newtheorem{definition}{Definition}
\newtheorem{observation}{Observation}
\newtheorem{fact}{Fact}

\begin{document}
%
\title{Testing Unsatisfiability of Constraint Satisfaction Problems via Tensor Products}

\author{Daya Ram Gaur \\
 Department of Math \& Computer Science \\ 
 University of Lethbridge \\
 Lethbridge, AB, Canada, T1K 3M4 \\
 Email: gaur@cs.uleth.ca \\
 http://www.cs.uleth.ca/$\sim$gaur
\And 
 Muhammad A. Khan\\
 InBridge Inc\\ 
 3582 30th Street N\\
 Lethbridge, AB, Canada, T1H 6Z4 \\
 Email: muhammad@inbridgeinc.com\\ 
 https://inbridgeinc.com/home-page/team/
}

\maketitle
 \begin{abstract}
We study the design of stochastic local search methods to prove unsatisfiability of a constraint satisfaction problem (CSP). For a binary CSP, such methods have been designed using the microstructure of the CSP. Here, we develop a method to decompose the microstructure into graph tensors. We show how to use the tensor decomposition to compute a proof of unsatisfiability efficiently and in parallel. We also offer substantial empirical evidence that our approach improves the praxis. For instance, one decomposition yields proofs of unsatisfiability in half the time without sacrificing the quality. Another decomposition is twenty times faster and effective three-tenths of the times compared to the prior method. Our method is applicable to arbitrary CSPs using the well known dual and hidden variable transformations from an arbitrary CSP to a binary CSP.
\end{abstract}


\section{Introduction}
Constraint satisfaction problems (CSPs) is a general and descriptive paradigm that is used to model and solve a variety of issues in diverse areas such as scheduling, planning, text analysis and logic. Satisfiability, graph homomorphism, answer set programming are all instances of CSPs. A CSP instance is a set of variables along with a set of relations on subsets of variables. A solution to a CSP satisfies all these relations. CSPs model NP-complete problems and are therefore computationally challenging to solve. One of the practical solution methods is backtracking coupled with constraint propagation. Constraint propagation aims at reducing the search space by enforcing a check on some necessary condition for a solution to exists. One could characterize methods, which use constraint propagation along with backtracking to solve a CSP, as ``complete methods" \citep{ghedira2013}. Of course, when a complete method fails to discover a solution to the CSP, we have a ``proof'' of the unsatisfiability. One can also use an ``incomplete method" such as local search to check for unsatisfiability of a CSP at each choice point in the search. 

Interest in incomplete methods for detecting unsatisfiability arises from the success of incomplete methods for solving problems; for example, walk-sat for satisfiability problem \citep{kautz2009}. A challenge to develop an efficient stochastic local search to determine unsatisfiability for propositional problems was issued by  \cite{selman1997computational}.
In the same year, \cite{gaur1997detecting} gave an incomplete method to determine the unsatisfiability of CSPs.  The last twenty years have seen a limited success in addressing the challenge due to \cite{selman1992new} in the context of CSPs. We surmise that there are primarily two reasons for this. The first reason has to do with the size of the microstructure, which is quadratic in the size of the CSP and can result in a computational blow-up. The second reason for limited use of these methods in backtracking is the difficulty in invoking the incomplete method at each choice point. This is due to the non-incremental nature of the technique, that is, one cannot reuse the work done in the prior stages.

In this paper, we exhibit another incomplete method using tensor product of graphs. The key idea is that the microstructure can be described as a union of graph tensor products. We relate the chromatic number of the microstructure to the chromatic numbers of the tensor products in the union. The decomposition of the microstructure into a union of graph tensors implies that upper bounds on the chromatic number of the microstructure can be computed more efficiently in practice. This is due to the facts that the tensor product graphs (i) are smaller than the microstructure, (ii) can be colored in parallel, and (iii) some very efficiently computable upper bounds on the chromatic number are known. Consequently, heuristic algorithms for coloring take significantly lesser time. The known upper bounds on the chromatic number  can be used at each choice point. The bookkeeping required for this is extremely fast. Our approach works for arbitrary $k$-ary CSPs with no restriction on the type of relations. 
 
In our approach, coloring a microstructure involves coloring a series of tensor graphs of the same total size as the microstructure. Each of of these computations can be performed in parallel. We give an infinite family of CSPs where the approach can establish unsatisfiability. Although, the use of efficiently computable upper bounds reduces the set of instances where the method can show unsatisfiability, the resulting speed-up is quite significant. 

We also report on experiments on random binary CSPs given by the not-all-equal relation. Our experiment design overcomes some of the difficulties \citep{achlioptas2001random} in the generation of unsolvable instances in \citep{gaur1997detecting}. We observe that a half of all the instances can be proved to be unsatisfiable using the tensor product decomposition for large $n$ (see Figure \ref{fig:n}). We also determine unsatisfiability using the original method of \cite{gaur1997detecting}, which is able to prove unsatisfiability of slightly less than half of the instances (see Figure \ref{fig:n}). Our method is almost twice as fast compared to the original method in general. For dense symmetric CSPs our method is twenty times faster, and is able to prove unsatisfiability of three-tenths of the instances compared to the original method (see Figure \ref{fig:p}).

In Section
\ref{sec:tensor}, 
we describe a way to decompose the microstructure of a binary CSP into graph tensors. We then relate the chromatic number of the microstructure to the chromatic numbers of constituent tensors. An important property of this decomposition is that the sizes of the individual tensors is smaller than size of the microstructure. Furthermore, the tensors can be colored in parallel and often very efficiently using  theoretical results on tensor products and graph coloring. A coloring of the microstructure can be computed very efficiently given a coloring of the tensors. Our approach works for arbitrary CSPs with arbitrary (symmetric or non-symmetric) relations. In Section \ref{sec:empirical} we give infinite families of CSPs for which unsatisfiability can be detected using the proposed tensor decomposition method. We evaluate our approach empirically and give an efficient method of generating unsatisfiable CSP instances across a spectrum. We observe that the tensor decomposition method proves unsatisfiability of more instances and is twice as fast compared to the original method of \cite{gaur1997detecting} on a comprehensive test set. For dense instances it is twenty times faster on average, though the fraction of instances proved unsatisfiable drops down. The tensor decomposition method also improves the incomplete method in \citep{benhamou2008new} as discussed in Section \ref{sec:lit}. Thus our approach partially answers  challenge 5 in \citep{selman1997computational}. 
Our tensor decomposition based framework can for  unsatisfiability of arbitrary CSPs, not just binary CSPs.

\section{Motivation}

Constraint solvers have enjoyed resounding success in finding solutions to large scale optimization problems with order of millions of variables. The two issues of `finding a solution if one exists' and `showing that there is no solution' are qualitatively very different. The former task is in NP and the latter is in Co-NP. 
The constraint satisfaction problems generated as shown below do not have a solution. These instances show that the state of art constraint solvers are woefully inadequate at answering the second question. Therefore new algorithms have to designed to show that a CSP does not have a solution.

The instances that we provide come from a conjecture due to  \cite{erdos1981}. The conjecture states that any union of $n$ cliques each of order $n$ such that no two cliques intersect in more than one vertex is $n$-colorable. The conjecture is still open. A natural question to ask is whether the number of cliques can be increased without increasing the chromatic number.  In fact there are graphs which are union of $n+1$ cliques each of order $n$ pairwise intersecting in at most one vertex which are not $n$-colorable.

Let \texttt{mat} be a $k+1 \times k$ matrix constructed as follows:

\begin{verbatim}
var elem = 1;
    for (j =1; j <= k; j++) {
        for (i = 1; i <= j ;i++) {
            mat[i][j] = elem;
            mat[j+1][i] = elem;
            elem = elem +1;
        }
    }
\end{verbatim}

The matrix for $n=8$ is

\begin{verbatim}
        [[1 2 4 7 11 16 22 29]
         [1 3 5 8 12 17 23 30]
         [2 3 6 9 13 18 24 31]
         [4 5 6 10 14 19 25 32]
         [7 8 9 10 15 20 26 33]
         [11 12 13 14 15 21 27 34]
         [16 17 18 19 20 21 28 35]
         [22 23 24 25 26 27 28 36]
         [29 30 31 32 33 34 35 36]]
\end{verbatim}

The rows of this matrix are the cliques, and any two cliques intersect in exactly one vertex. The number of cliques is one more than the size of each clique.  For each even $n=4,6,8,\ldots$ the corresponding graph is not $n$-colorable.  It is worthwhile to note that for odd $n$, the resulting graph is $n$ colorable. The coloring constraints can be described using pairwise compatibility relations or using $n$-ary relations. It has been argued that the $n$-ary allDifferent constraint \citep{regin1994filtering} is a better way to model. An allDifferent constraint on $n$ variables with at most $n-1$ different domain values can be never be satisfied (Hall's Matching Theorem). Therefore, the use of allDifferent constraint in the model easily shows that the graph is not $k$ colorable for all $k \le n-1$.

We use the allDifferent constraint (one for each clique) to model the coloring problem (with $n$ colors) and solve it using a state of art constraint solver  (CP optimizer in IBM ILOG CPLEX 12.9), default options with a parallel search using 32 threads on Intel(R) Xeon(R) CPU E5-2683 v4 @ 2.10GHz. For $n=4,6$ the IBM ILOG CPLEX solver took 0.190, 6.570 seconds respectively. For $n=8$ the program did not finish within 45 minutes.  

For $n=6$, the number of branches was 2,023,823 with 996,880 fails.  The search speed (number branches/second) was 876,113.9. The search grows exponentially and it would impossible for the current techniques to prune and show that union of $n+1$ cliques of size $n$ each (as constructed above) is not $n$-colorable for even $n\ge 8$.

\section{Related Research}\label{sec:lit}

Several constraint propagation methods: arc-consistency, path-\break consistency and $k$-ary consistency (weak and strong) \citep{montanari1974networks,mackworth1985complexity} are in wide spread use. The idea is to infer additional constraints and eventually derive a CSP that is quickly shown to be unsolvable.   \cite{regin1994filtering,regin1996generalized} developed constraint propagation methods for all-different constraints.  \cite{van1989constraint} studied bound constraints and cardinality constraints in constraint logic programming. 

Various subclasses of CSP are known to be solvable in polynomial time.  \cite{freuder1982sufficient} was the first to relate the structure of the CSP to its complexity and showed that a tree structured CSP can be solved in polynomial time using arc-consistency.   \cite{dechter1989tree} introduced the notion of induced-width and related it to the level of consistency required for a backtrack-free search.  \cite{gottlob2014treewidth} proposed to view the structure of a CSP as a hypergraph, and proved that a CSP with constant hyper-width is solvable in polynomial time. This result subsumes all the above mentioned results relating structure to complexity. For a 
comprehensive survey on the complexity issues in CSP, see the paper by \cite{carbonnel2016tractability}.

\cite{schaefer1978complexity} dichotomy theorem states that every family of satisfiability instances is either in polynomial time or is NP-complete. \cite{hell1990complexity} proved a similar dichotomy theorem for graph homomorphism. Both, satisfiability and graph homomorphism are CSPs. Therefore it is natural to ask the question for CSPs in general.  \cite{feder1998computational} conjectured the existence of a dichotomy theorem for CSPs. \cite{bulatov2017dichotomy} recently proved, using algebraic methods, that every family of a non-uniform CSP is either solvable in polynomial time or is NP-complete, thereby establishing the Feder--Vardi conjecture.

The microstructure is a Karp reduction from CSP to $k$-clique. Therefore, graph-theoretic properties can be used to determine the solvability of a CSP. For instance, if the microstructure is perfect then the CSP can be solved in polynomial time \citep{salamon2008perfect}. Recently, a very interesting approach based on patterns, forbidden in the microstructure has been developed. Cooper, Jeavons and Salamon defined the broken triangle property (BTP) in a microstructure with respect to a given variable ordering \citep{cooper2010generalizing}. They showed that if the microstructure is BTP free with respect to some ordering then the CSP can be solved in polynomial time. Furthermore, the existence of a BTP free ordering can also be determined in polynomial time. Tractability results for several other forbidden patterns have been established by \cite{cohen2012tractability}. A dichotomy is not known for the BTP free property. Hence, several relaxations of the BTP free property have been studied. In particular,  m-fBTP \citep{el2017btp,el2018new} implies tractability using arc-consistency or arc-consistency with forward checking.

Nearly all of the incomplete methods that test for unsatisfiability, rely on  the microstructure \citep{jegou1993decomposition}. The  incomplete method of \cite{gaur1997detecting} is to color the microstructure  of the CSP using a prescribed number of colors. They showed that the method is intrinsically different from methods based on arc-consistency and also performed a limited empirical evaluation. \cite{bes2005proving} performed a detailed assessment and argued for the limited applicability of the technique.  \cite{benhamou2008new}, extended the approach of  \cite{gaur1997detecting} to propose a new incomplete method based on the notion of dominance in CSPs and established wide applicability. We comment on the relationship of our work to the work of  \cite{benhamou2008new}. They integrated the approach of \cite{gaur1997detecting} with generalized arc consistency (GAC) on all-different constraint due to \cite{regin1994filtering}. Either the original CSP is shown to be unsatisfiable using the method of \cite{gaur1997detecting} or a new  CSP (a single all-different constraint) is formed using the original CSP and its coloring. GAC on the new CSP may reduce the domains in the original CSP. If that happens, a new microstructure is created and the process iterates. Greedily coloring the microstructure (Step 5, in Algorithm 1) is a crucial component in \citep{benhamou2008new}. Any improvement to Step 5, improves their algorithm overall. Our decomposition provides a faster way to color the microstructure, thereby improving Step 5 of Algorithm 1 in \citep{benhamou2008new}.


\section{Decomposition into Tensors}\label{sec:tensor}

We give the definitions and an overview of our approach in this section. The technical details appear in the next Section. 
A binary CSP is a 3-tuple $(X, D, {\cal{R}})$, where $X$ is the set of variables, $D$ is the set of domain values, and ${\cal{R}}$ is a set of binary relations, called constraints, of pairwise compatible values for pairs of variables in $X$. Without loss of generality, we assume that each variable $x\in X$ takes values in the domain $D$. The set of pairwise compatible values for variables $x,y$ are specified by a relation $R_{xy} \subseteq D \times D$. The relation $R_{xy}$ is {\it symmetric} if $(a,b) \in R_{xy}$ then $(b,a) \in R_{xy}$. Given an assignment $x=a , y=b$, constraint $R_{xy}$ is {\it satisfied} if  $(a, b) \in R_{xy}$. A CSP is said to be {\it satisfiable} if every variable is assigned a value such that all the constraints are satisfied. If no such assignment of values to the variables exists, then the CSP is {\it unsatisfiable}. Associated with each CSP is a graph $G$ called the {\it constraint graph}, in which the nodes are the variables, and $(x,y)$ is an edge if there is a relation $R_{xy} \in {\cal{R}}$. 

The following reduction due to \citep{dechter1989tree,rossi1990equivalence} can be used to transform a non-binary CSP with $n$ constraints into a binary CSP. For each allowed $k$-tuple of values $(v_1, v_2, \ldots , v_k)$ in a $k$-ary relation $R(x_1, x_2, \ldots, x_k)$; we have a node (value). An edge connects two $k$-tuples belonging to different relations $R_i$, $R_j$ if the "underlying assignment" is compatible. The nodes that belong to the $k$-tuples from the same constraint form an independent set. Each constraint corresponds to a variable, the $k$-tuples of values are the values that are allowed by the constraint, in the new binary CSP. The compatible nodes assign the same value to the common variables in the original CSP.  A clique in the microstructure of the binary CSP constructed as above corresponds to a solution of the original CSP. If the microstructure can be colored with $< n$ colors, then the non-binary CSP is not satisfiable. There is another reduction method due to  \cite{peirce1931collected}, which proves that binary CSPs have the same expressive power as arbitrary CSPs. Please see \citep{rossi1990equivalence} for further details of the reductions. Therefore, from now we will assume without loss of generality that the CSP is binary.

\begin{definition}[Microstructure]
Consider a CSP $(X, D, {\cal{R}})$ with constraint graph $G$ and   relations in ${\cal{R}}$. The {\it microstructure} $G_\mu$ of  the CSP \citep{jegou1993decomposition} is a graph defined as follows: For every unconstrained pair of variables $x, y$, assume that $R_{xy}\in {\cal{R}}$ is the universal relation allowing all pairs of values for $x, y$. There is a node $x_v$ for every pair of variable $x\in X$ and value $v\in D$.  There is an edge $(x_u, y_w)$ if and only if $x\neq y$ and $(u,w) \in R_{xy}$. 
\end{definition}

The following observation can be used to test for the unsatisfiability of a CSP.

\begin{observation}\citep{gaur1997detecting}
An $n$-variables CSP is unsatisfiable if the microstructure is $(n-1)$-colorable.
\end{observation}

Binary relations  can be visualized as directed  graphs possibly with loops. If the relation is symmetric we replace the pair of oppositely directed edges by an undirected edge. The not-all-equal relation ($\neq$) over domain $\{1,2,3\}$, is the symmetric relation $\neq_{xy}=\{(1,2),(1,3),(2,3),(2,1),(3,1),(3,2)\}$, represented by a triangle graph. The equal ($=$) relation over domain $\{1,2,3\}$, is $=_{xy}=\{(1,1),(2,2),\break (3,3)\}$, represented by a three-node graph with only edges being loops on the nodes. Let $C_k$ be the complete graph with loops on every node. This corresponds to a complete relation on $k$ values (every pair of values is compatible with the relation).  Let $N_k$ be the complete graph without any loops. This corresponds to the not-all-equal relation over $k$ values. Given a graph $G$ with $k$ nodes, let $G'$ denote the {\it complement} of $G$.
Finally, the chromatic number of a graph $G$ denoted $\chi(G)$ is the least number of colors needed to color the nodes of $G$ such that all pairs of nodes that share an edge, have different colors.

 \begin{figure*}[htp]
 \centering
 \begin{minipage}{.5\textwidth}
   \centering
   \includegraphics[width=.8\linewidth]{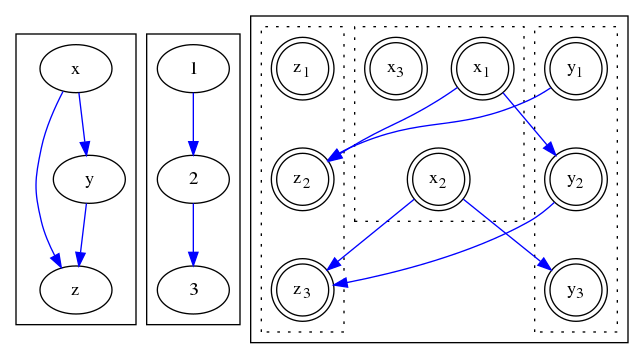}
   \caption{Tensor product ($\otimes$) of digraphs}
   \label{fig:test1}
 \end{minipage}%
 \begin{minipage}{.5\textwidth}
   \centering
   \includegraphics[width=.7\linewidth]{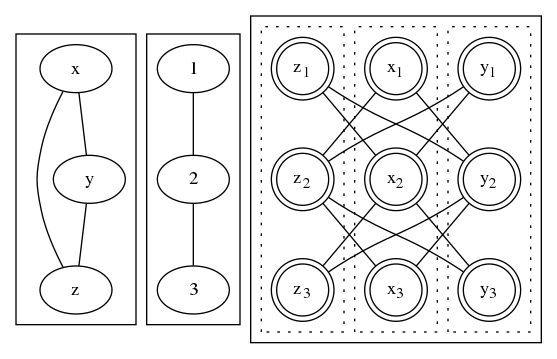}
   \caption{Tensor product ($\otimes$) of graphs}
   \label{fig:test2}
 \end{minipage}
 \vspace*{-6mm} 
 \end{figure*}

\subsection{Tensor Product of Digraphs} 
A {\it digraph} (or directed graph) $G$ with node set (or vertex set) $V(G)$ and edges set $E(G)$ is a graph with each edge endowed with a direction. The (directed) edges of a digraph are typically called {\it arcs}. Here, we make no notational distinction between an edges joining the nodes $a$ and $b$ and an arc directed from $a$ to $b$, denoting both by $(a, b)$. The {\it degree} ${\rm deg}_{G}(v)$ of a node $v$ in a digraph (graph) $G$ is the total number of arcs (edges) meeting the node with loops at $v$ counted twice. We simply write ${\rm deg}(v)$ if $G$ is clear from the context.   

A digraph $G$ is {\it symmetric} if $(a, b)\in E(G)$ implies $(b, a)\in E(G)$. A symmetric digraph can be represented by an undirected graph by replacing every pair of symmetric arcs by a single undirected edge. In this way, graphs can be considered special instances of digraphs. The digraphs (graphs) we consider have no parallel arcs (edges), that is, more than one arcs (edges) with the same initial and terminal node (joining the same pair of nodes), but may possess loops. We refer to \cite{chartrand2015} for the standard terminology concerning graphs and digraphs. It is easy to observe that any binary relation on a nonempty set can be expressed as a digraph and vice versa. The {\it underlying graph} of a digraph is the graph obtained by dropping all the arc directions. The {\it chromatic number of a loopless digraph} is defined to be the chromatic number of its underlying graph.  

Introduced as an operation on binary relations in \citep{whitehead1912}, the tensor product can be naturally defined for digraphs (and hence for graphs) as follows \citep{hammack2016}. Given digraphs (resp. graphs) $G$ and $H$, the tensor product $G\otimes H$ has node set $V(G)\times V(H) =\{a_b : a\in V(G), b\in V(H)\}$, the Cartesian product of $V(G)$ and $V(H)$, with an arc directed from $a_b$ to $c_d$ (resp. an edges joining $a_b$ and $c_d$) if $(a, c)\in E(G)$ and $(b, d)\in E(H)$.

\begin{figure*}[h]
    \centering
 \includegraphics[width=\textwidth]{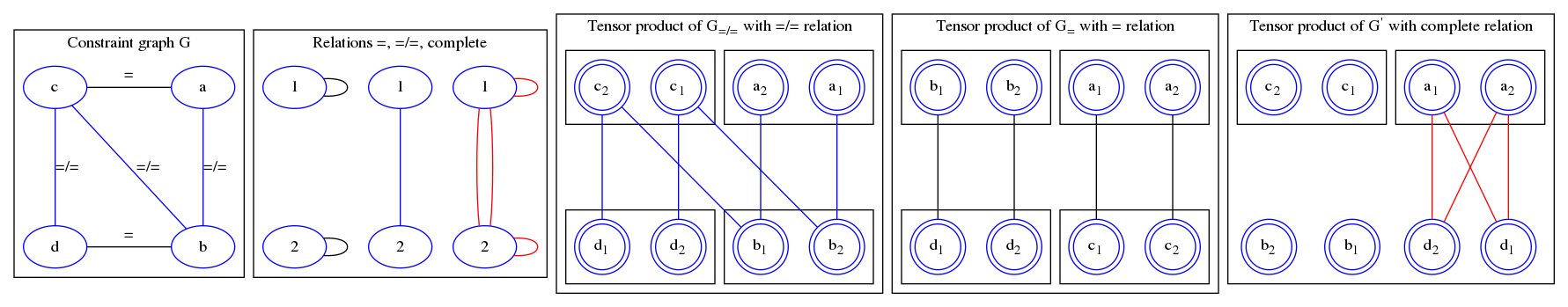}
   \caption{Leftmost graph is the constraint graph $G$ of a CSP with constraints $R_=$, $R_{\neq}$. The subgraph of $G$ in black (blue) is $G_=$ ($G_{\neq}$). The second box contains the three graphs corresponding to relations $=, \neq, C_2$. The third graph is the tensor product $G_{\neq} \otimes R_{\neq}$. The fourth graph is the tensor product  $G_= \otimes R_{=}$. The last graph is the tensor product  $G' \otimes C_2$.   The labels within () on the nodes correspond to colors.}
  \label{fig:csp}
  \vspace*{-4mm} 
  \end{figure*}

Clearly, $G\otimes H = H\otimes G$. Also, $|V(G\otimes H)|=|V(G)|\cdot|V(H)|$, where $|S|$ denotes the size of a set $S$. Moreover, for any $a_b \in V(G\otimes H)$, ${\rm deg}(a_b) = {\rm deg}(a)\cdot{\rm deg}(b)$. For any digraphs (or graphs) $G$ and $H$, $G \otimes H$ contains $|V(H)|$ copies of $G$ and $|V(G)|$ copies of $H$. 
Figure \ref{fig:test1} shows the tensor product of two digraphs and Figure \ref{fig:test2} shows that tensor product of their underlying graphs. We observe that the latter contains twice as many edges as the former. A CSP, along with the decomposition of the microstructure into constituent tensors is shown in Figure \ref{fig:csp}.

Note that $G\otimes H$ is loopless irrespective of $G$ or $H$ having loops or not, and therefore, $\chi(G\otimes H)$ is well-defined. If $G$ and $H$ are loopless digraphs, then $\chi(G)$ and $\chi(H)$ are also well-defined. Since each copy of $G$ in $G\otimes H$ can be properly colored using $\chi(G)$ colors, we have the following well-known property of tensor products \citep{hammack2016}.   
\begin{fact}
If $G$ and $H$ are loopless digraphs (or graphs), then 
\begin{equation}\label{eq:bound}
\chi(G \otimes H) \le \min\{\chi(G), \chi(H)\}.
\end{equation}
\end{fact}

When $G$ and $H$ are loopless graphs  \citep{hedetniemi1966} conjectured the stronger relation 
\begin{equation}\label{eq:hedetniemi}
\chi(G \otimes H) = \min\{\chi(G), \chi(H)\}.
\end{equation}

The digraph analogue of the conjecture does not hold \citep{poljak1981}. Hedetniemi's conjecture has been one of the most important open problems on graph coloring. It is known to be true in some cases including if one of the graphs is $4$-colorable or the complete graph $N_k$ \citep{hammack2016}. However, the conjecture has been disproved in general recently \citep{shitov2019}.

\subsection{Decomposition of microstructure}
The microstructure graph $G_\mu$ of a binary $k$-CSP $(X, D, {\cal{R}})$ with constraint graph $G$ and $|D|=k$ can also be viewed as a digraph as follows:

\begin{definition}[Microstructure - the general case]
The digraph $G_\mu$ has a node $x_v$ for every pair of variable $x\in X$ and value $v\in D$.  There is an arc $(x_u, y_w)$ if and only if $x\neq y$ and $(u,w)\in R_{xy}$. As before, for every unconstrained pair of variables $x, y$, we assume that $R_{xy}\in {\cal{R}}$ is the universal relation allowing all pairs of values for $x, y$.
\end{definition}

Given $x, y\in X$, we denote by $G_{xy}$ the digraph obtained by deleting all arcs from $G$ except $(x, y)$. Recall, that $R_{xy}$ is the relation that lists compatible values for the variables $x,y$. Comparing the definitions of microstructure and tensor product, we have the following. 
\begin{observation}
$\displaystyle G_{\mu} = \hspace*{-4mm} \bigcup_{(x, y)\in E(G)} \hspace*{-4mm}(G_{xy}\otimes R_{xy})\ \ \bigcup \ \ (G'\otimes C_k ).$
\end{observation}

If every  $R_{xy}$ is the same relation $R$, one gets the simpler expression 
$$G_{\mu} = (G\otimes R) \cup (G'\otimes C_k ).$$

Using the fact that the chromatic number of a union of digraphs is bounded above by the product of their chromatic numbers, we get 
\[\displaystyle \chi(G_{\mu}) \le \hspace*{-4mm}\prod_{(x, y)\in E(G)} \hspace*{-4mm}\chi(G_{xy}\otimes R_{xy})\cdot \chi(G'\otimes C_k ),
\]
which reduces to 
\[\chi(G_{\mu}) \le \chi(G\otimes R) \cdot \chi(G'\otimes C_k ),
\]
if the relations $R_{xy}$ are all the same. 

Now, relations (\ref{eq:bound}) and (\ref{eq:hedetniemi}) can be used to upper bound $\chi(G_{xy}\otimes R_{xy})$ or $\chi(G\otimes R)$, but not $\chi(G'\otimes C_k)$ as $C_k$ has loops. However, note that $G'\otimes C_k$ is a graph of significantly smaller number of edges than $G_{\mu}$. We therefore, have  
\[\displaystyle \chi(G_{\mu}) \le \hspace*{-3mm}\prod_{(x, y)\in E(G)} \hspace*{-4mm}\min \{\chi(G_{xy}), \chi(R_{xy})\}\cdot \chi(G'\otimes C_k ),
\]
or the reduced form 
\[\chi(G_{\mu}) \le \min\{\chi(G), \chi(R)\} \cdot \chi(G'\otimes C_k ).
\]
Since $G_{xy}$, $G$, $R_{xy}$ and $G'\otimes C_k$ are all considerably smaller sized graphs than $G_\mu$, we can obtain an upper bound on $\chi(G_\mu)$ rather efficiently.

\subsection{Loopless $k$-CSP}
If all the relations, considered as digraphs, are loopless in a CSP, then we can expedite the computation of the upper bound. We use the following observation regarding the tensor decomposition and the inequality (\ref{eq:bound}). If a relation (equivalently a digraph) $R$ is partitioned (edges and not the nodes in the digraph are partitioned) into relations $R_a$ and $R_b$, then for any digraph $G$, 
$$G \otimes R = (G \otimes R_a) \cup (G \otimes R_b),$$ and
$$\chi(G \otimes R) \le \chi(G \otimes R_a) \cdot \chi(G \otimes R_b).$$ 

 We partition the complete relation as $C_k = I_k \cup N_k$, where $I_k$ is the `$=$' relation, and $N_k$ is the `$\neq$' relation. For a  $k$-CSP $(X, D, {\cal{R}})$ with all the relations in ${\cal{R}}$ the same loopless relation $R$, and constraint graph $G$, the decomposition (\ref{eq:bound}) requires us to compute $\chi(G \otimes R) \cdot \chi(G' \otimes C_k)$, which by the observation above is $\le \chi(G \otimes R) \cdot\chi(G' \otimes I_k) \cdot \chi(G' \otimes N_k)$. As $R$ and $N_k$ are loopless, we know by (\ref{eq:bound}) that $\chi(G \otimes R) \le \min\{\chi(G),\chi(R)\}\le \chi(R)\le k$. Moreover, since (\ref{eq:hedetniemi}) holds when one of the graphs is $N_k$ (complete graph), $\chi(G \otimes N_k) = \min\{\chi(G),\chi(N_k)\}\le k$. The tensor product $G' \otimes I_k$ is $k$ copies of $G'$, which can be colored by $\chi(G')$ colors. Hence, the following observation. 

\begin{observation} \label{obsv:fast}
If for a loopless $k$-CSP on $n$ variables, $k^2 \cdot \chi(G') < n$ then the CSP is unsatisfiable. Note that this only requires us to color the complement of the constraint graph resulting in a substantial speed-up.
\end{observation}

As there are infinitely many graphs with a constant chromatic number $k$ such that $k^3 < n$, we have an infinite family for which this decomposition method is able to prove unsatisfiability.

\section{Experimental Results}\label{sec:empirical}
 
\cite{gaur1997detecting} generated random CSPs using the method of  \cite{smith2001constructing}. They then identified unsolvable CSPs using an ILP solver. This set is used for testing. Some of the unsolvable instances were arc-consistent, and others were not. The method limits them to small-sized instances ($n=10$) as the solvability of the CSP needs to be determined. Furthermore, the relation between a pair of variables is asymmetric with a high probability. We conduct our experimental evaluation on symmetric CSPs resulting in symmetric microstructure graphs. The instances are arc-consistent, unsolvable, and of large size. All experiments were implemented on Intel(R) Core(TM) i5-4210U CPU @ 1.70GHz with 8GB of RAM in julia 0.6.2.

We generate an \cite{erdos1960evolution} random graph $G_{n,p}$ over $n$ nodes with an edge probability $p$. For a constant and fixed $p$, the family of graphs is known as {\it dense}.  The dense graphs have interesting global properties. For instance, the clique number (the maximum number of pairwise adjacent nodes) of $G_{n,1/2}$ is $2\log{n}$, and the maximum degree is $n/2$ almost always, and far away from the chromatic number of $G_{n,1/2}$, which is close to $n / (2\log{n})$ almost always \citep{bollobas1988chromatic}. Thus, for any $k < n / 2\log{n}$, the $G_{n,1/2}$ is not $k$-colorable. This gives us an easy way to obtain  unsolvable symmetric $k$-CSPs, avoiding the use of an ILP solver as in \citep{gaur1997detecting}. Recall that a CSP is arc-consistent if, for any pair of variables $(x,y)$, for all values in the domain of $x$, there is some consistent value in the domain of $y$ (and vice-versa). The resulting instances are therefore arc-consistent for all $k \ge 2$. 

We do not use the method known as Model B \citep{smith2001constructing} for generation of random CSPs. Model B was studied in   \citep{achlioptas2001random}. They showed that the model is flawed is the sense that the random CSP instances do not have asymptotic threshold and almost all instances they are over-constrained. Unsatisfiability of these can be readily determined using trivial local inconsistencies.
We wanted to avoid unsatisfiability detection by an easy use of local consistency checking. Therefore we used Erdos-Reyni random graphs to highlight cases where local consistency fails, but our method succeeds in practice. 

We consider the loopless symmetric $k$-CSP corresponding to the `$\neq$' relation with constraint graph $G_{n,p}$. The microstructure $G_\mu = G_{n,p} \otimes N_k \cup G'_{n,p} \otimes C_k$ is the union of the two tensor products. The unsatisfiability of such a CSP can be proved in three ways. 

The pseudo-code for the three methods below is listed in the appendix. 

\begin{itemize}
\item $\mu$ (microstructure) method:
This is the original method of \cite{gaur1997detecting}. We color the microstructure $G_\mu$ and check if $\chi( G_\mu) < n$. If so, the CSP is unsolvable.

\item $\otimes$ (tensor decomposition) method:
This is one of the decomposition methods developed in this paper. If $\chi(G_{n,p}\otimes N_k) \cdot \chi(G'_{n,p} \otimes C_k)  < n$ then the CSP $G_{n,p}$ is unsolvable. Since, equation (\ref{eq:hedetniemi}) is satisfied when one of the graphs is $N_k$, we have $\chi(G_{n,p} \otimes N_k) = k$. To determine the unsatisfiability using the second method we only color $\chi(G'_{n,p} \otimes C_k)$. 

\item $\otimes_E$ (fast tensor decomposition) method: Observation \ref{obsv:fast} gives a third way.  We obtain a coloring of the complement of the constraint graph and if $k^2 \chi(G') < n$, then the CSP is unsatisfiable. 
\end{itemize}

We use the following greedy algorithm as a baseline to color a graph. We color a node using the first available color, given an ordering of the nodes. If no color is available, then we increase the size of the palette. We consider six orderings; the nodes are ordered in the decreasing order of degree, and five random ones. One can replace this step with a method that chooses the next node to color dynamically, such as Brelaz's method. This replacement can only improve the results.

We generated $G_{n,p}$ instances for $n$ in the interval $[10,100]$ in steps of $10$ and $p$ in $[0.1,1]$ in steps of $0.1$. We used the above three methods to color the microstructure. For each $p$, $1620$ instances were generated. The number of instances in the increasing order of $n$ are $300,600,900,1200,1500,1800,2100,2400,2700,2700$. The number of instances in the increasing order of $k \ge 2$ are $3000,2700,2400,2100,1800,1500,1200, 900,600$.

We first present the aggregate results. Next, we examine the effect of varying the parameters. Finally, we describe the running times. Table \ref{table:aggregate} lists the number of instances proved unsatisfiable by a subset of methods. 
A total of $9456$ instances of $16200$ were proved unsatisfiable using one of the methods. The $\mu$, $\otimes$, $\otimes_E$ methods proved unsatisfiability of $8974$,$9312$,$3216$ instances respectively. The decomposition based methods proved unsatisfiability of $482$ additional instances which were not detected by the old method. The old method detected $144$ additional instances than the other methods.

\begin{table}
\begin{center}
\begin{tabular}{|cccc|}
\hline
\hline
$\mu$ & $\otimes$ & $\otimes_E$ & \# \\\hline\hline
 false      & false      & false           & 6744 \\
true       & false      & false           &144  \\
true       & true       & false           & 5614 \\
true       & true       & true            &3216 \\
false      & true       &false           & 482 \\
\hline
\end{tabular}
\caption{Number of Unsolvable instances $G_{n,p}$}
\label{table:aggregate}
 \end{center}
\end{table}

\begin{table*}[htp]
\begin{center}
\begin{tabular}{|cccc|cccc|cccc|}
\hline 
$n$ & $\mu$& $\otimes$ & $\otimes_E$ & $p$ & $\mu$ & $\otimes$ & $\otimes_E$ & $k$ & $\mu$ & $\otimes$& $\otimes_E$  \\\hline 
\hline 
10   & 247  & 196  & 68 & 0.1 & 216  & 163 & X&  X & X & X & X \\
20  & 460  & 430 &  145 &0.2 & 283  & 261 &  X &2  & 2887 & 2768 & 1632\\
30  & 615  & 616 &  237 & 0.3 & 416  & 451 & X & 3 & 1964 & 2028 & 600\\
40  & 753  & 770  &  278 &0.4 & 533  & 593  & 66 & 4  & 1348 & 1459  & 264\\
50  & 872  & 932  &  330 & 0.5 & 717  & 755  & 192 & 5  & 953 & 1004  & 210 \\
60  & 1011 & 1059  &  362 & 0.6 & 915  & 997 &  237 & 6  & 674  & 747  & 180\\
70  & 1141 & 1186   & 408 & 0.7 & 1170 & 1289 & 262 & 7  & 473  & 541  & 150 \\
80  & 1230 & 1295  &  419 & 0.8 & 1484 & 1563 &  368 & 8  & 343  & 372 & 120 \\
90  & 1297 & 1387 & 472 & 0.9 & 1620 & 1620  & 561 & 9  & 212  & 258   & 60\\
100 & 1348 & 1441  &  497 & 1.0 & 1620 & 1530  & 1530 & 10 & 120  & 135 & X\\\hline
\end{tabular}
\caption{Number of provably unsatisfiable instances grouped by $n,k,p$. Columns $\mu$ ($\otimes$)[$\otimes_E$] is the number of instances proved unsolvable by coloring the microstructure (tensor product of $G'$ with complete relation $C_k$)[coloring $G'$]}
\label{table:2}
\end{center}
\end{table*}

\begin{figure}[!ht]
  \caption{Relative Performance w.r.t. $p$}
  \centering
\includegraphics[height=2in]{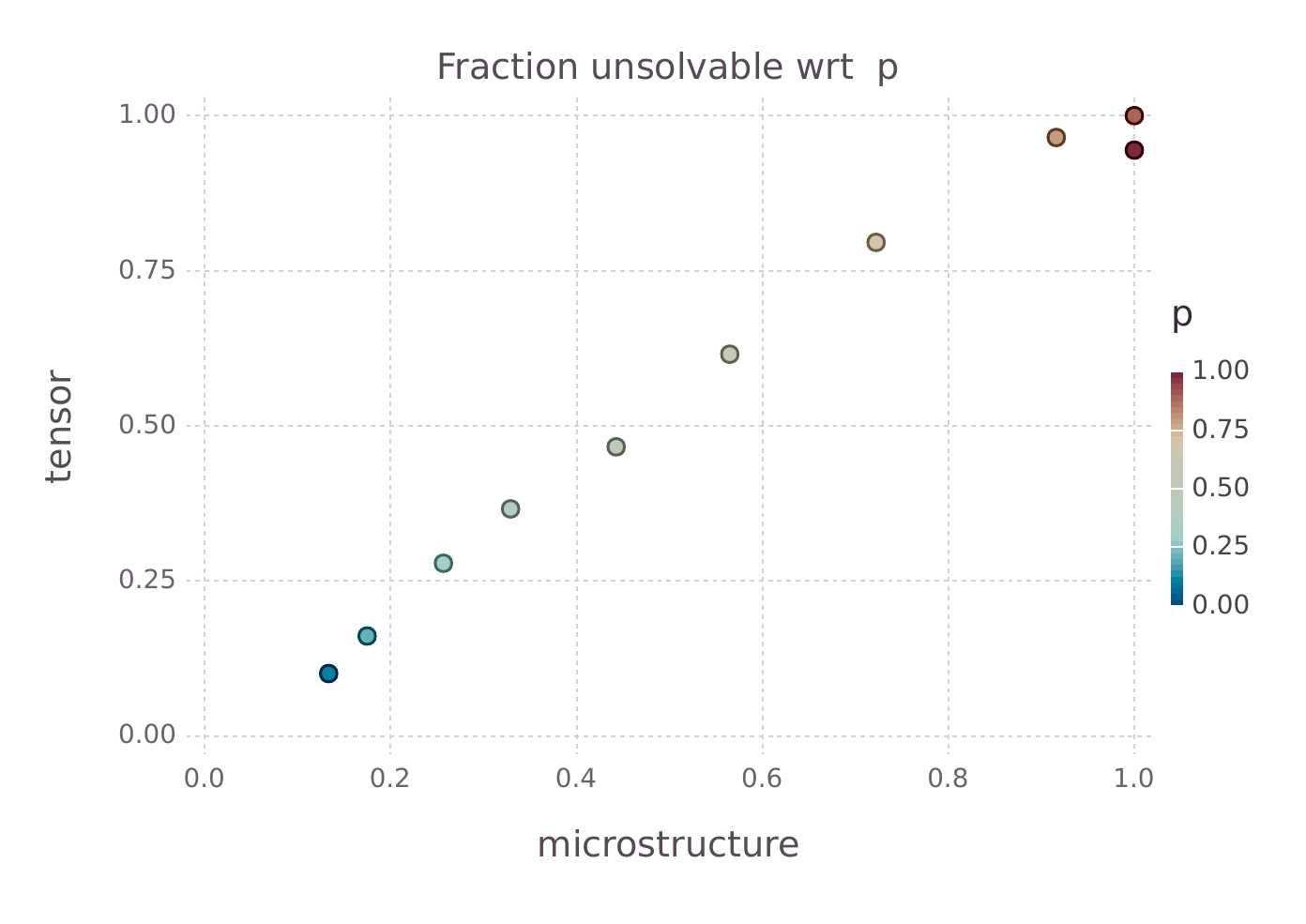}
\label{fig:p}
\vspace*{-8mm} 
 \end{figure}

\begin{figure}[!ht]
  \caption{Relative Performance w.r.t. $k$}
  \centering
 \includegraphics[height=2in]{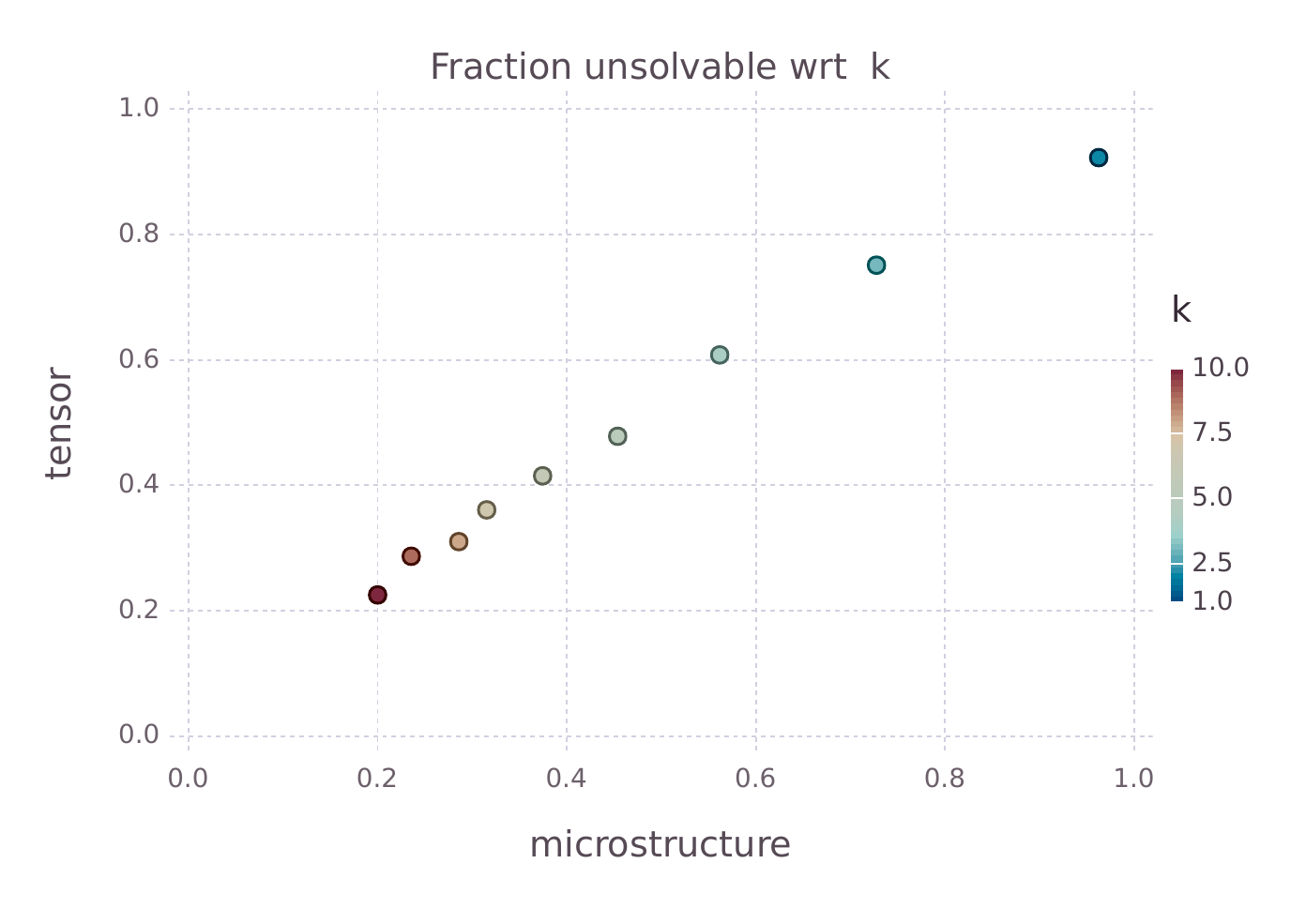}
 \label{fig:k}
 \vspace*{-6mm} 
  \end{figure}

\begin{figure}[!ht]
  \caption{Relative Performance w.r.t. $n$}
  \centering
 \includegraphics[height=2in]{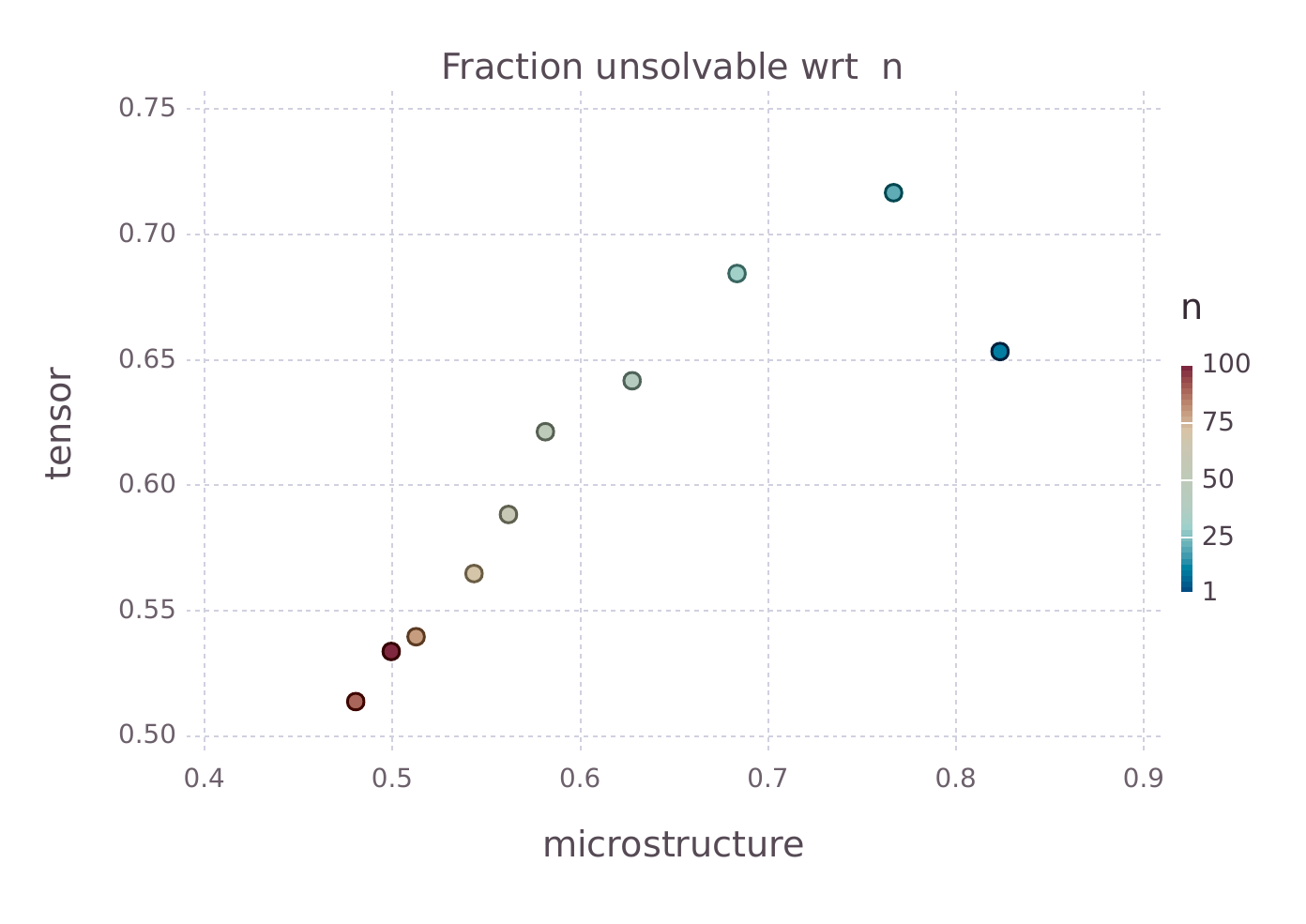}
 \label{fig:n}
 \vspace*{-6mm} 
    \end{figure}

\begin{figure}[!ht]
  \caption{Relative Performance of $x=\otimes$ and $y=\otimes_E$}
  \centering
 \includegraphics[height=2in]{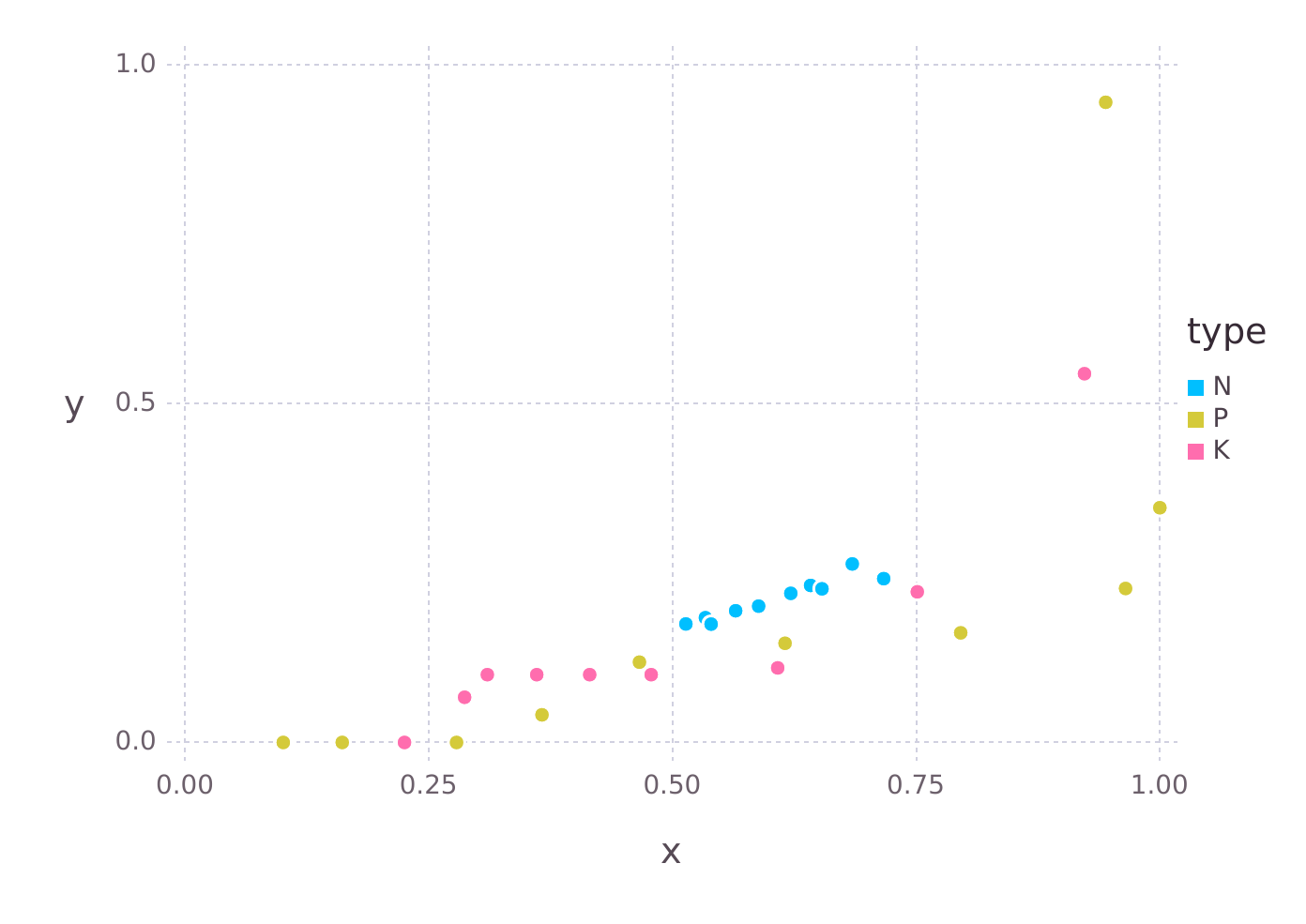}
 \label{fig:newplot}
 \vspace*{-8mm} 
    \end{figure}

The number of instances proved unsolvable by the three methods when grouped by $n,p,k$ are shown in the Table \ref{table:2}. 'X' indicates missing values.
For each grouping by $n,p,k$ we determine the fraction of the total instances that are proved unsolvable by the first two strategies ($\mu, \otimes$). The three plots (Figures \ref{fig:p}, \ref{fig:k}, \ref{fig:n}) give the relative performance of the two strategies. The fraction of the instances determined unsolvable by coloring the microstructure are represented on the $x$-axis, whereas the $y$-axis represents the fraction of the instances with a proof of unsatisfiability obtained by coloring the tensor product.  
Figure \ref{fig:newplot} show the relative performance of the faster tensor decomposition based strategy $\otimes_E$ compared to the $\otimes$ strategy grouped by $n,p,k$. The percentage of instances proved unsatisfiable by $\otimes$ ($\otimes_E$) strategies are on the $x$-axis ($y$-axis).

\vspace*{-4mm} 
 
\begin{figure}[!ht]
  \caption{Runtime}
  \centering
\includegraphics[height=1.8in]{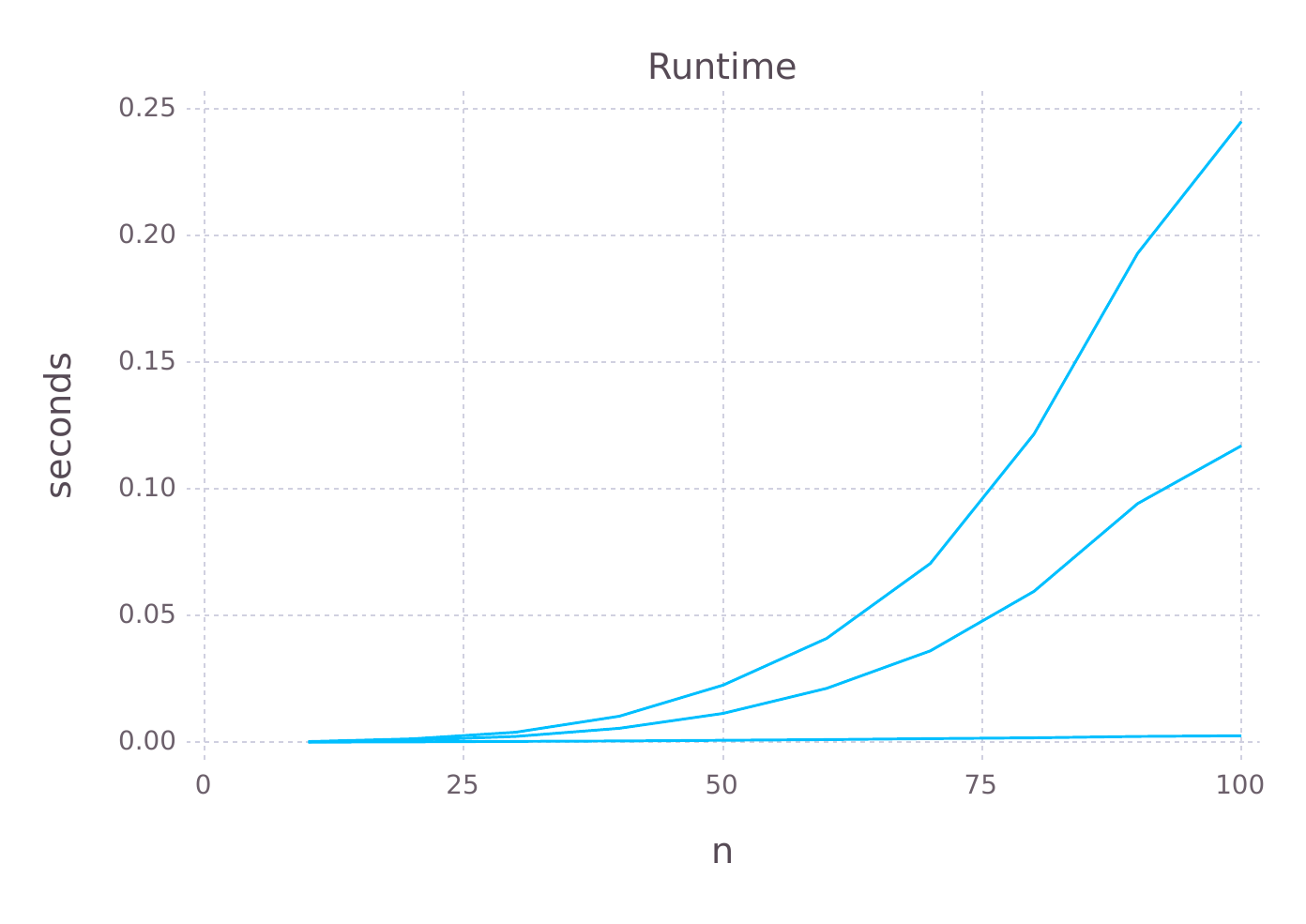}
\label{fig:runtime}
\vspace*{-8mm} 
 \end{figure}

Finally, a comment on the running times. 
 If both $G \otimes N_k$ and $G' \otimes C_k$ are to be colored, then they can be colored in parallel, thereby reducing the time. The earlier method \citep{gaur1997detecting} is inherently sequential. The average times as a function of $n$ is shown in Figure \ref{fig:runtime}. The curves in the order from the highest to the lowest are for $\mu$, $\otimes$, $\otimes_E$ respectively. The proposed method $\otimes$ is twice as a fast as the old method $\mu$. The faster tensor method $\otimes_E$ is faster by a factor of $40$ for values of $p >0.75$. The $\otimes_E$ method is about $20$ times faster on average (over all values of $p$),  and can prove unsatisfiability of 20\% of all the instances considered. 

\section{Discussion}

Let us call a CSP, a $2$-CSP if all the variable domains are of size $2$. In our experiments, we observe that the relative fraction of provably unsolvable $2$-CSP instances is high ($>0.55$) for all the methods. A coloring $2$-CSP can be decided in polynomial time. We suspect, the randomized coloring method proves unsatisfiability, almost always for $2$-CSPs. A proof of this would be serious progress towards answering challenge 5 in \citep{selman1997computational}. 

The tensor decomposition methods are applicable for any CSP with an arbitrary number of different  $k$-ary relations on pairs of variables. In our experiments, we assumed that all the binary relations are the same and symmetric (not-all-equal). It would be interesting to conduct further experiments in the generalized setting. If there are $r$ different relations, possibly asymmetric, then there are $r+1$ tensor products that have to be colored. They can be colored in parallel, reducing the running time by a factor of $O(1/(r+1))$. The upper bound on the chromatic number of the microstructure is computed as a product of the chromatic numbers of the $r+1$ tensor products. A theoretical analysis, establishing that $r=1$ is the worst-case (or prove otherwise) for the tensor decomposition method is the second interesting question. Some immediate progress can be made here by assuming that $m <r$ relations give rise to loopless digraphs, in which case we can use Observation \ref{obsv:fast}, and bound the chromatic number by the number of nodes in the component tensor. 

Third, and very interesting question is whether the product upper bound on the chromatic number in the decomposition can be strengthened. This requires the development of new heuristic ways of combining the colorings of the tensor products. Any progress on this question would immediately increase the efficacy of our method. This in turn would further speedup the method of \cite{benhamou2008new}.
We end with a note. In our illustrations the tensor decomposition is based on the relations in the CSP. However, this does not have to be the case, the decomposition is not fixed. In fact, a single CSP relation can be decomposed into multiple relations, (and multiple CSP relations can be combined into one).  This raises the possibility of the development of other general methods for decomposing the microstructure graphs into their constituent tensors. 

\emph{Acknowledgements:} The authors thank the referees for the comments.

\section{Appendix}

This section lists the pseudo-code for the three methods $\mu, \otimes, \otimes_E$. We assume the existence of a method \verb|color(G)|, which colors a graph $G$ greedily, and returns the number of colors used. We also assume existence of the tensor product operator ($\otimes$). julia language provides the $\verb|kron|$ operator which can be used to construct the adjacency  matrix of $G \otimes R$, given the adjacency matrices for $G, R$. 

We use the following convention in the listings below. $G$ is the constraint graph. $R_1, R_2, \ldots, R_m$ are the relation graphs on the edges of $G$. $G(R_i)$ is the graph obtained using the subset of edges in $G$ constrained by relation $R_i$. The variables have the same domain $D$ and the number of values in $D$ is $k$.

\begin{algorithm}
\caption{$\mu$ method \citep{gaur1997detecting}}\label{alg:mu}
\begin{algorithmic}[1]
\Require Binary CSP, given by $G$ and $R_i's$.
\Procedure{$\mu$}{$G, R_1, R_2, \ldots, R_m$}
\State $G_\mu = \phi$ \Comment{Initialize $G_\mu$ to an empty graph.} 
\For {$i \in [1..m]$} \Comment{Compute the microstructure}
\State $G_\mu =  G_\mu \cup (G(R_i) \otimes R_i)$
\EndFor
\State $G_\mu = G_\mu \cup (G' \otimes C_k)$ 
\State \Return \verb|color|($G_\mu$) \Comment{Color the microstructure}
\EndProcedure
\end{algorithmic}
\end{algorithm}

Algorithm \ref{alg:mu} is the method of \cite{gaur1997detecting}. The pseudo-code in Algorithm \ref{alg:tensor} describes the tensor decomposition based method. We illustrate the parallelism while coloring the tensors in the decomposition using the \verb|@parallel| construct in julia. This executes the for loop in parallel and the results of individual computations are combined (reduced) using the * operator.

\begin{algorithm}
\caption{$\otimes$ method}\label{alg:tensor}
\begin{algorithmic}[1]
\Require Binary CSP, given by $G$ and $R_i's$.
\Procedure{$\otimes$}{$G, R_1, R_2, \ldots, R_m$}
\State colors = 1 
\State colors = \verb|@parallel (*)| \Comment{Execute the for loop in parallel and reduce the results using the (*) operator}
\For {$i \in [1..m]$} 
\If {$R_i$ is loopless} \State
    \Return $\min\{|G(R_i)|, |R_i|\}$ \Comment{Using Eq (\ref{eq:bound})}
    \Else \State
    \Return \verb|color|($G\otimes R_i$)
\EndIf
\EndFor
\State \Return colors * \verb|color|($G'\otimes C_k$)
\EndProcedure
\end{algorithmic}
\end{algorithm}

Finally, we describe the method based on Observation \ref{obsv:fast}. This method works only when all the relations are loopless. It colors just the complement of the constraint graph. The number of colors needed for the tensors in the decomposition are estimated using (\ref{eq:bound}).

\begin{algorithm}[H]
\caption{$\otimes_E$ method}\label{alg:fasttensor}
\begin{algorithmic}[1]
\Require Binary CSP, given by $G$ and $R_i's$. $R_i's$ are loopless. 
\Procedure{$\otimes$}{$G, R_1, R_2, \ldots, R_m$}
\State colors = 1 
\For {$i \in [1..m]$} 
\State colors = colors * $\min\{|G(R_i)|, |D|\}$ \Comment{O(1) computation.}
\EndFor
\State \Comment{Decompose $C_k$ as $I_k \cup N_k$}
\State \Return colors * $|D|$ * \verb|color|($G'$)
\EndProcedure
\end{algorithmic}
\end{algorithm}

\begin{small}
\bibliographystyle{aaai}
\bibliography{csp.bib}
\end{small}


\end{document}